\documentclass{interact}

\usepackage{xurl}
\usepackage{url}
\usepackage{hyperref}
\usepackage{pdflscape}
\usepackage{bm}

\usepackage[capitalize]{cleveref}
\crefname{section}{Sec.}{Secs.}
\Crefname{section}{Section}{Sections}
\Crefname{table}{Table}{Tables}
\crefname{table}{Tab.}{Tabs.}

\usepackage{natbib}% Citation support using natbib.sty
\bibpunct[, ]{(}{)}{;}{a}{}{,}% Citation support using natbib.sty
\renewcommand\bibfont{\fontsize{10}{12}\selectfont}% Bibliography support using natbib.sty

\theoremstyle{plain}% Theorem-like structures provided by amsthm.sty

\theoremstyle{definition}

\theoremstyle{remark}

\begin{document}

%\articletype{ARTICLE TEMPLATE}

\title{Deep Clustering of Remote Sensing Scenes through Heterogeneous Transfer Learning}

\author{
\name{Isaac Ray\textsuperscript{a}\thanks{CONTACT Isaac Ray Email: null@stat.tamu.edu} and Alexei Skurikhin\textsuperscript{b}}
\affil{\textsuperscript{a}Department of Statistics, Texas A\&M University, College Station, TX, USA; \textsuperscript{b}Intelligence and Space Research, Los Alamos National Laboratory, NM, USA}
}

\maketitle

\begin{abstract}
This paper proposes a method for unsupervised whole-image clustering of a target dataset of remote sensing scenes with no labels. The method consists of three main steps: (1) finetuning a pretrained deep neural network (DINOv2) on a labelled source remote sensing imagery dataset and using it to extract a feature vector from each image in the target dataset, (2) reducing the dimension of these deep features via manifold projection into a low-dimensional Euclidean space, and (3) clustering the embedded features using a Bayesian nonparametric technique to infer the number and membership of clusters simultaneously. The method takes advantage of heterogeneous transfer learning to cluster unseen data with different feature and label distributions. We demonstrate the performance of this approach outperforming state-of-the-art zero-shot classification methods on several remote sensing scene classification datasets.
\end{abstract}

\begin{abbreviations}
DNN - Deep Neural Network;
GMM - Gaussian Mixture Model;
BNP - Bayesian Nonparameteric;
DPGMM - Dirichlet Process Gaussian Mixture Model;
UMAP - Uniform Manifold Approximation and Projection;
ViT - Vision Transformer;
VI - Variational Inference
\end{abbreviations}

\textbf{Funding Details} The work of Isaac Ray was partially supported by the NSF under Grant 2139772.

%\begin{keywords}
%Deep Learning; Bayesian Clustering; Manifold Projection; Land Use Classification; Transfer Learning
%\end{keywords}

\section{Introduction}
\label{sec:intro}
Labelled data are expensive and time consuming to obtain, whereas unlabelled data are abundant. This is especially true in the remote sensing domain, where tremendous amounts of unlabelled aerial and satellite images are collected daily \citep{national_noaa_nodate, crosier_how_2022}. 
%Remote sensing plays a vital role in various applications, including environmental monitoring and disaster management. 
To address this, unsupervised clustering methods aim to group whole images together without an explicit labelling scheme. These clusters can then be used in downstream research tasks such as latent feature discovery and data collection in fields including land use analysis \citep{6946898}, climate science \citep{zscheischler_climate_2012}, and urban planning \citep{rahman_classification_2019}. 

Since images are too complex to cluster directly, we would like to reduce the feature space by first learning deep and semantically meaningful features from the data \citep{ren_deep_2022} and then projecting the learnt feature space into a lower dimension to have a useful notion of distance between sparse feature groups \citep{steinbach2004challenges}. Although clustering methods can then group similar scenes based on how far apart their deep features are, most traditional clustering methods such as K-Means \citep{lloyd_least_1982}, Gaussian Mixture Models (GMM), and Density-based Spatial Clustering of Applications with Noise (DBSCAN) \citep{ester_density-based_1996} rely on specifying a certain number of clusters or neighbours a priori. This structure is too rigid for applications where feature discovery is a goal, or when there is no domain-informed choice for these parameters, leading to suboptimal model selection methods such as ‘elbow’ finding for scree plots \citep{milligan_examination_1985}. And, while they may not need a lower-dimensional feature space, many current deep clustering techniques require training an expensive deep neural network clustering layer for every target dataset.

To remedy these problems, we propose a heterogeneous transfer learning approach \citep{zhuang_comprehensive_2019} to deep image clustering. We first finetune a pretrained deep neural network (DNN) on a source set of labelled remote sensing images whose feature and label spaces may be different from our target images' \citep{day_survey_2017}. We claim that this finetuning only needs to be done once to generalise across multiple target image datasets. Using this DNN, we extract a high-dimensional deep feature vector for each unlabelled target image. We then use a manifold projection technique to embed these high-dimensional deep features into a low-dimensional Euclidean space. Finally, we adopt a Bayesian Nonparametric (BNP) approach to clustering with the Dirichlet Process Gaussian Mixture Model (DPGMM) \citep{blei_variational_2006} so that we can infer both the number of clusters and cluster membership from these embedded features.

This paper addresses the challenge of clustering remote sensing scenes without relying on supervision or cluster specification. We summarise our contributions as follows: (1) We propose a highly flexible pipeline for nonparametric image clustering using a combination of a pretrained and finetuned DNN, manifold projection, and the DPGMM.
(2) We apply our method to a variety of land use datasets and demonstrate its superior performance compared to state-of-the-art unsupervised methods in remote sensing.

\begin{figure*}[ht!]
  \centering
   \includegraphics[width=0.9\linewidth]{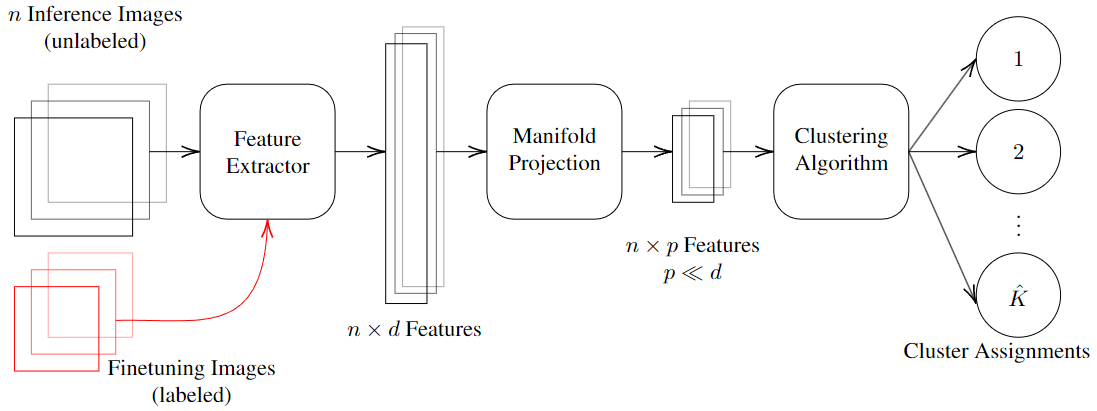}
   %\resizebox{0.95\columnwidth}{!}{\input{figures/pipeline_fig_tikz}}

   \caption{Diagram of the proposed image clustering pipeline: given $n$ unlabelled inference images to cluster, we extract $d$-dimensional features from a DNN trained on a general image dataset and finetuned on a remote sensing dataset. We then project these features onto a much lower $p$-dimensional approximated manifold and apply our clustering algorithm to these lower dimensional features.}
   \label{fig:pipeline}
\end{figure*}
\section{Related Works}
\label{sec:related}

Before detailing our nonparametric image clustering approach, we briefly review the existing literature on deep feature extraction using DNNs, deep clustering methods for images, heterogeneous transfer learning using DNNs, image clustering in remote sensing, and applications of BNP techniques in deep learning. We note that as a general framework for clustering, our proposed clustering pipeline shown in \cref{fig:pipeline} and described in \cref{sec:method} encompasses many different approaches described in the literature. For example, \citet{mcconville_n2d_2019} used an autoencoder as a feature extractor, the Uniform Manifold Approximation and Projection (UMAP) algorithm \citep{mcinnes_umap_2018} as a manifold projection, and the GMM as a clustering method. However, they do not examine transfer learning applications and assume that the true number of clusters is known.

\subsection{Deep Clustering}

DNN approaches for unsupervised image clustering have been advancing rapidly in the past few years. Within-image clustering (also called segmentation) adopts similar DNN-based practices for feature extraction, but for a fundamentally different task \citep{saha_unsupervised_2022, orbanz_nonparametric_2008}. \citet{radford_learning_2021} discuss transfer learning using Natural Language Processing (NLP) as a form of semi-supervision and provide an overview of ‘zero-shot’ learning as an alternative to clustering, which we compare with in our experiments. We categorise deep image clustering approaches into those that are end-to-end and learn their deep feature representations and cluster memberships simultaneously, and those which are multistep and apply a clustering method to separately extracted features. Most of the methods in the former category require the true number of clusters to be specified a priori \citep{zhou_deep_2022, adaloglou_exploring_2023, caron_deep_2018, qian_stable_2023}. Some recent BNP approaches for end-to-end deep clustering and feature extraction do not make this assumption. This includes methods in which feature extractors such as variational autoencoders \citep{bing_diva_2023, lim_deep_2021} and convolutional neural networks \citep{wang_dnb_2022} are trained simultaneously with a BNP clustering method to jointly learn feature representation, the number of clusters, and cluster membership. However, these approaches come with a high computational cost and do not take advantage of transfer learning to avoid expensive DNN training epochs for each target dataset.

Similarly to our method, a variety of multistep approaches to clustering with deep features have been explored. \citet{zhou_deep_2022} propose a method for direct clustering of deep features extracted similarly to what we propose. They also examine why K-Means \citep{lloyd_least_1982} is suboptimal for clustering these features. However, they still rely on the true number of clusters being specified and found that finetuning was counterproductive, which we do not observe in our work. In contrast, \citet{adaloglou_exploring_2023} demonstrated strong transfer learning performance when clustering with transformer models using self-supervised pretraining, although they also use the true number of labels. Methods which do not rely on the true number of clustering include DeepDPM, a model proposed by \citet{ronen_deepdpm_2022} that uses BNP in the form of a modified split/merge Expectation Maximisation (EM) algorithm \citep{jain_splitting_2007} to estimate both the cluster membership and count using pseudo-probabilities estimated with neural networks. They found that deep clustering methods such as SCAN \citep{van_gansbeke_scan_2020} are often very sensitive to the mispecification of the number of clusters, leading to suboptimal performance. However, their algorithm is complex and requires substantial engineering efforts to implement and tune.

For more details on deep clustering methods, we refer the reader to the comprehensive survey on deep clustering by \citet{ren_deep_2022} and note that our method falls into their class of DNN based transfer learning approaches.

\subsection{Remote Sensing}

Several previous works have addressed the problem of transfer learning and clustering for remote sensing image analysis. \citet{petrovska_deep_2020} propose a method that follows our clustering framework, consisting of feature extraction using a pretrained CNN, manifold projection using principal component analysis, and clustering using support vector machines. However, they only considered two remote sensing datasets and did not leverage recent advances in DNN architectures, self-supervised learning, or BNP techniques. \citet{dimitrovski_aitlas_2023, dimitrovski_current_2023} presented AiTLAS, a framework for analysing the impact of different feature extraction backbones, pretraining strategies, and finetuning datasets on the performance of transfer learning for remote sensing image classification. Their results showed that Vision Transformers (ViT) pretrained on ImageNet-21k \citep{deng_imagenet_2009} and finetuned on a large-scale remote sensing dataset exhibited good transfer learning performance across multiple target datasets. \citet{ayush_geography-aware_2020, cong_satmae_2022} both combined spatio-temporal information with contrastive self-supervised learning to improve the representation learning of remote sensing images when considering image segmentation tasks. They demonstrated that their methods outperformed conventional self-supervised learning methods that only used pixel-level information. However, all of these methods rely on the true number of clusters being specified. 

Only a few BNP approaches have been explored in remote sensing and most have been used for image segmentation on data such as hyperspectral imagery \citep{mantripragada_evaluation_2022} or panchromatic satellite images \citep{shu_object-based_2015}. To the best of our knowledge, our work is the first to use DPGMM clustering of features extracted using DNN-based transfer learning on a large number of real-world satellite and aerial imagery datasets.

\section{Method}
\label{sec:method}

In this section, we describe the proposed method for clustering remote sensing scenes without labels or supervision on a target dataset, summarised in \cref{fig:pipeline}. The method consists of three main steps: (1) finetuning a general pretrained deep neural network on a domain specific imagery dataset and extracting a feature vector from each image in an unlabelled target imagery dataset in the same domain using the finetuned model, (2) reducing the dimension of the feature vector by embedding it into a lower dimensional space while preserving structural relationships between features, and (3) clustering the embedded features using Bayesian nonparametrics to infer the number and membership of clusters simultaneously. Under the framework laid out by \citet{zhuang_comprehensive_2019} and investigated by \citet{day_survey_2017}, our proposed method can be categorised as a heterogeneous transfer learning approach which does not require labelling of the target dataset but can benefit from labelled source data. 

\begin{figure*}[tbh]
  \centering
   \includegraphics[width=0.95\linewidth]{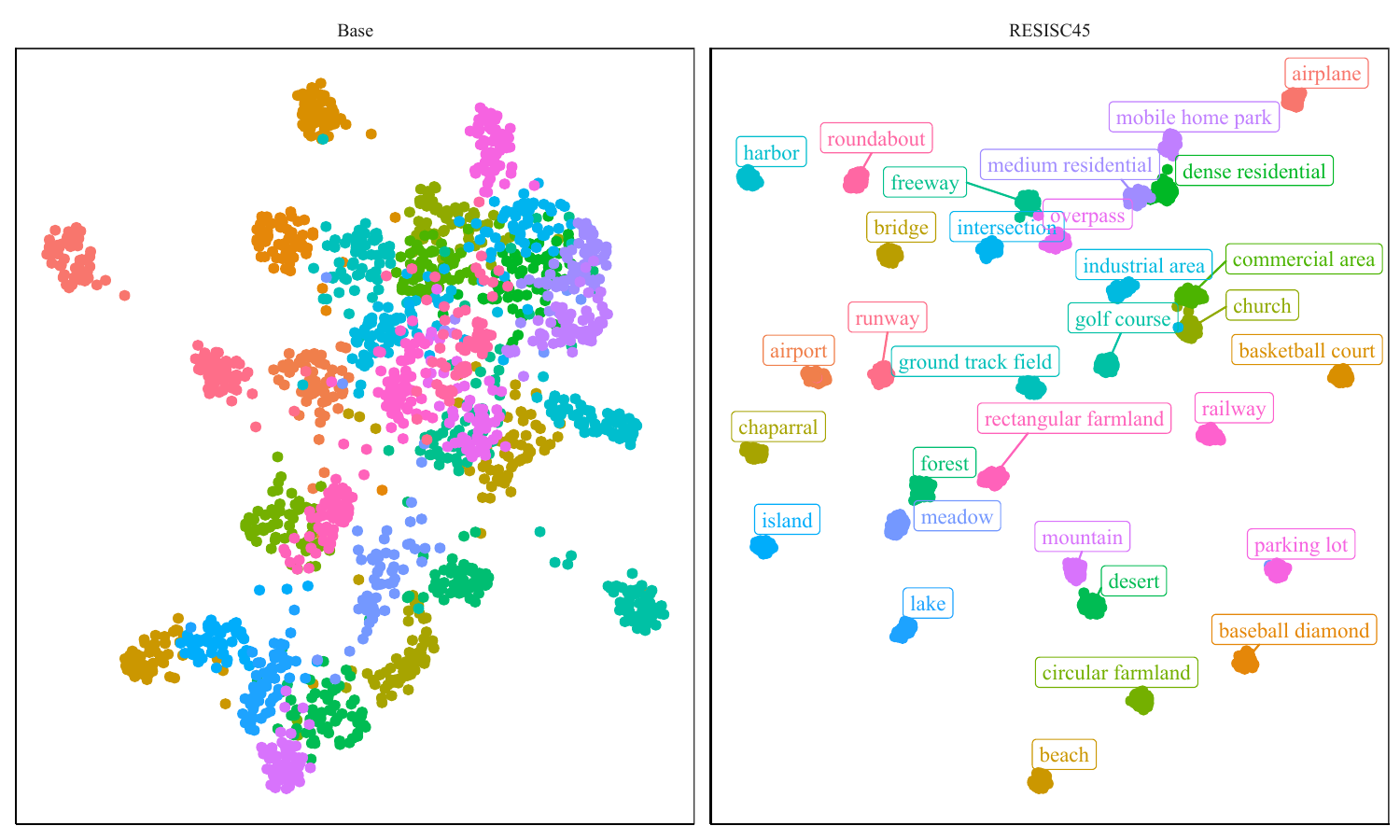}
   \caption{Visualisation of the embedded features for the Optimal-31 dataset when using the base DINOv2-L model as a feature extractor (left), and when using the DINOv2-L model finetuned on the RESISC45 dataset (right). The colours and labels denote the ground truth labels. As verified in \cref{tbl:main}, the features from the finetuned model are much easier to cluster.}
   \label{fig:optimal31-ft}
\end{figure*}

\subsection{Choosing a Feature Extractor} We use a vision transformer model (ViT) as the feature extractor, which is a neural network that applies the transformer architecture \citep{vaswani_attention_2017} to images. ViT models have been shown to achieve state-of-the-art results in various computer vision tasks, such as image classification, object detection, and semantic segmentation \citep{dosovitskiy_image_2020}. However, ViT models typically require a large amount of labelled data to be trained from scratch, which is not available in our setting. Therefore, we use a ViT model called DINOv2 that has been pretrained on a large-scale image dataset using a discriminative self-supervised learning method. \citet{oquab_dinov2_2023} showed that DINOv2 learns robust visual features without supervision by encouraging the agreement between multiple views of the same image. Furthermore, they showed that these models often outperform other self-supervised methods and even some supervised methods on various downstream tasks. Similarly, in our preliminary exploration we found that this choice of backbone dominated other architecture and training choices, such as various pretrained CNN models and ViT models from scratch. We denote a DINOv2 model using a ViT-Base/14 Vision Transformer backbone \citep{dosovitskiy_image_2020} as DINOv2-B, and similarly DINOv2-L when using a ViT-Large/14 backbone. Models were obtained through the PyTorch Hub \citep{paszke_pytorch_2019}. We describe these models without finetuning as (Base) feature extractors. When finetuning or inference is performed, we resize images to match the expected input size for the feature extractor. The features we extract are the values of the final pooling layer with sizes $d = 768$ for DINOv2-B and $d = 1024$ for DINOv2-L.

\subsection{Finetuning} We use the DINOv2 model as a pretrained feature extractor and finetune it on a labelled source dataset of remote sensing images. The target dataset is an unlabelled dataset of remote sensing images that we want to cluster. As part of our heterogeneous transfer learning approach, it can have a different distribution or resolution than the source dataset, as we do not assume any prior knowledge about it. The source dataset could be any existing remote sensing scene classification dataset, but motivated by \citet{dimitrovski_current_2023} and our own exploratory analysis, we focus on the NWPU-RESISC45 dataset \citep{cheng_remote_2017} due to its good generalisation across a wide variety of other remote sensing datasets when used in combination with a Vision Transformer-based classifier. We also observe in \cref{tbl:satindata} that its images are quite heterogeneous with 45 different ground truth labels and a wide range of ground sampling distances, which we believe contribute to its transferability to other land use datasets. The source dataset does not need to have the same distribution or label space as the target dataset, as we only use it to adapt the DINOv2 model to the remote sensing domain. After finetuning the DINOv2 model on the source dataset, we use it to extract features from the target dataset. 

We denote the DINOv2 models finetuned on RESISC45 as (RESISC45) feature extractors. We use the finetuned DINOv2 model to process each image in the target dataset into a $d$-dimensional feature vector from the last hidden pooling layer of the model. Because the DNN was pretrained and finetuned, these feature vectors represent high-level and semantic information in the image. However, the feature vectors are still too high-dimensional to be clustered effectively using off-the-shelf distance-dependent methods due to the curse of dimensionality \citep{steinbach2004challenges}. Therefore, we use a manifold projection technique to reduce the dimension of the feature vectors by embedding them into a lower-dimensional space that preserves the local structure of the features.

\subsection{Choosing a Manifold Projection} Similar to \citet{mcconville_n2d_2019}, we use the UMAP algorithm \citep{mcinnes_umap_2018} to reduce the dimension of our features while preserving local structure. While we explored various manifold projection techniques such as t-SNE \citep{maaten_visualizing_2008}, TriMap \citep{amid_trimap_2022}, and PaCMAP \citep{wang_understanding_2020}, we found that the choice of algorithm had little effect on the final clustering metrics. We chose UMAP because its hyperparameters did not strongly influence clustering performance and it scales well to very large datasets. 

UMAP is based on the assumptions that the data are uniformly distributed on a locally connected Riemannian manifold and that the Riemannian metric is approximately locally constant. The algorithm first constructs a weighted k-nearest neighbours graph from the data, representing its fuzzy topological structure. It then finds a $p$ dimensional embedding with a similar fuzzy topological structure by using stochastic gradient descent to minimise cross-entropy loss. We apply UMAP to the $d$-dimensional feature vectors extracted from the target dataset and obtain a $p$-dimensional embedded feature vector for each image, where $p \ll d$. The embedded feature vectors are in a Euclidean space that approximates the manifold structure of the original feature space.

\subsection{Choosing a Clustering Algorithm} The final step of the method is to cluster the embedded feature vectors. Since our feature vectors are embedded in a low-dimensional Euclidean space, we can use distance-based methods under the assumption that points which are close together in the embedded space likely belong to the same cluster. We elect to use BNP methods, which do not impose a fixed parametric form on the data distribution, but rather allow the complexity of the model to grow with the data. This is especially useful for clustering problems, where the number of clusters $\hat{K}$ is unknown and may vary depending on the data. 

% All equations must be numbered
% Equations must be referred to as 'Equation (1)' or 'Equations (2)-(4)'
% Every variable/function/term must be defined at first use ever if it is well-known or has an implied meaning
% All variables should be represented by a single letter/character only
% Scalars are italic (default), vectors are bold italic, matrices/tensors are bold
% 
One of the most prominent BNP models is the Dirichlet process (DP) \citep{ferguson_bayesian_1973}, which is a prior distribution over discrete probability measures. The DP can be used to model the uncertainty about the functional form of a distribution over parameters in a model. The DP is specified by a base distribution and a concentration parameter. We denote a distribution $G$ that follows a DP with base distribution $G_0$ and concentration $\alpha$ as $G \sim DP(G_0, \alpha)$. The base distribution $G_0$ is the expected value of the DP, whereas the concentration parameter controls the variability of the DP around the base distribution. The DP has the property of being discrete with probability one, which means that it can be used to model an infinite number of clusters.

%Describe the Dirichlet Process (based on stick breaking)
Based on this definition, the DP would serve as a prior over our multinomial mixing distribution. However, since our goal is clustering, we are more interested in the posterior distribution of the cluster labels which follow this mixing distribution. We choose to conceptualise the DP as a stick-breaking process \citep{sethuraman1994constructive} so we can more explicitly characterise the distribution of our cluster labels. In this construction, the cluster labels $\boldsymbol{z}$ follow a multinomial distribution whose probabilities $\boldsymbol{\pi}$ are defined recursively as $\pi_k = \phi_k\prod_{j=1}^{k-1}(1-\phi_j)$, where $\boldsymbol{\phi}$ are i.i.d. Beta random variables $\boldsymbol{\phi}\overset{i.i.d.}{\sim} \mathcal{B}(1, \alpha)$ whose shape is controlled by the concentration $\alpha$. In the standard stick-breaking construction, we allow $k\rightarrow \infty$ such that the $\pi_k$ values tend to zero. However, in practice we will truncate the maximum number of clusters at $\overline{K}$ by setting $\phi_{\overline{K}} = 1$ and $\pi_k = 0$ when $k > \overline{K}$. This allows us to have a computationally efficient algorithm while still inferring $\hat{K}\in\{1,\dots,\overline{K}\}$.

%Describe the Dirichlet Process Mixture Model (Dirichlet Process as prior over 
%We use the DP as a prior distribution for the mixing distribution in a Gaussian mixture model (GMM). A GMM is a probabilistic model that assumes our embedded feature vectors are generated from a mixture of multivariate Gaussian distributions, each with unknown parameters. 
Now that we have a constructive definition for cluster membership under our DP prior, we choose to model the individual clusters of our mixture distribution as being multivariate Gaussian distributions; also known as the Gaussian mixture model (GMM). For each of the $k \in \{1,\dots,\hat{K}\}$ Gaussian clusters, we need to infer the distribution's mean $\boldsymbol{\mu}_k$ and precision $\boldsymbol{\Sigma}_k$. To make inference computationally tractable, we choose to put a conjugate priors on our distributional parameters. We place a normal prior on the means with prior precision $\gamma$; $\mathcal{N}(\boldsymbol{0}, \gamma^{-1}{\textbf I})$, and a Wishart prior on the precisions with $\nu$ degrees of freedom and a  prior scale matrix ${\textbf D}$; $\mathcal{W}(\nu, {\textbf D})$.

By combining the DP prior with the GMM, we have the Dirichlet Process Gaussian Mixture Model (DPGMM) \citep{Rasmussen1999TheIG}. The full hierarchical specification of the DPGMM that we will use is given by:
\begin{align}
  \phi_k &\sim \mathcal{B}(1, \alpha) \\
  \pi_k &= \phi_k\prod_{j=1}^{k-1}(1-\phi_j) \\
  \boldsymbol{\mu}_k &\sim \mathcal{N}(\boldsymbol{0}, \gamma^{-1}{\textbf I}) \\
  \boldsymbol{\Sigma}_k &\sim \mathcal{W}(\nu, {\textbf D}) \\
  \boldsymbol{z} | \boldsymbol{\pi} &\sim \textrm{Multinomial}(\boldsymbol{\pi}) \\
  \boldsymbol{y} | \boldsymbol{z}, \boldsymbol{\mu}, \boldsymbol{\Sigma} &\sim \mathcal{N}(\boldsymbol{\mu}_{\boldsymbol{z}}, \boldsymbol{\Sigma}^{-1}_{\boldsymbol{z}})
\end{align}

Although a variety of strategies have been introduced to sample from or estimate the posterior of this model, we choose to use a variational inference (VI) algorithm to approximate the posterior distribution of our cluster membership $\boldsymbol{z}$ given the embeddings $\boldsymbol{y}$. In particular, we employ the mean-field assumption: given a family of distributions $q_\Theta(\boldsymbol{z})$ parameterised by variational parameters $\Theta$, we approximate our desired posterior $p(\boldsymbol{z}|\boldsymbol{y},-)$ by minimising the Kullback-Leibler (KL) divergence between them over $\Theta$ \citep{bishop_pattern_2006_dpgmmvi}. 

Because we chose conjugate priors for our latent variables 
$\{\boldsymbol{\phi}, \boldsymbol{\mu}, \boldsymbol{\Sigma}, \boldsymbol{z}\}$, 
their conditional distributions are Beta, Normal-Wishart, and Multinomial which all belong to the exponential family. By choosing our variational family to be a product of corresponding exponential family distributions, we can derive coordinate ascent updates for our variational parameters $\Theta$ in closed form. The closed form expressions for the coordinate ascent updates under this paradigm are given by \citet{blei_variational_2006}.

\section{Experiments}
\label{sec:experiment}

We begin by providing additional details about the implementation of our clustering pipeline described in \cref{sec:method}. Next, we present our main results for our method's performance on a wide range of different remote sensing datasets when using different choices of feature extractor in \cref{tbl:main} and compare against current state-of-the-art zero-shot classification models in \cref{tbl:zeroshot-compare}. Finally, we examine the performance impact of using non-BNP clustering approaches on remote sensing datasets.

%\begin{landscape}
\setcounter{table}{1}
\begin{sidewaystable}
%\begin{table}
%\small
\centering
\setlength\tabcolsep{2pt}
\begin{tabular*}{\linewidth}{@{\extracolsep{\fill}} lllllllllllll}
\toprule
\multicolumn{1}{c}{Feature Extractor} & \multicolumn{3}{c}{DINOv2-B (Base)} & \multicolumn{3}{c}{DINOv2-L (Base)} & \multicolumn{3}{c}{DINOv2-B (RESISC45)} & \multicolumn{3}{c}{DINOv2-L (RESISC45)} \\
\cmidrule(l{3pt}r{3pt}){2-4} \cmidrule(l{3pt}r{3pt}){5-7} \cmidrule(l{3pt}r{3pt}){8-10} \cmidrule(l{3pt}r{3pt}){11-13}
Dataset & ACC & NMI & ARI & ACC & NMI & ARI & ACC & NMI & ARI & ACC & NMI & ARI\\
\midrule
RESISC45* & .60 (.01) & .70 (.00) & .44 (.01) & \textbf{.72} (.01) & \textbf{.78} (.00) & \textbf{.59} (.01) & - & - & - & - & - & -\\
CLRS & .56 (.01) & .61 (.00) & .39 (.01) & .64 (.01) & .68 (.00) & .48 (.01) & .78 (.01) & .75 (.00) & .66 (.01) & \textbf{.79} (.01) & \textbf{.78} (.00) & \textbf{.68} (.01)\\
EuroSAT & .32 (.01) & .56 (.00) & .29 (.01) & .36 (.01) & .59 (.00) & .32 (.01) & \textbf{.38} (.02) & \textbf{.60} (.00) & \textbf{.33} (.01) & .35 (.01) & .58 (.01) & .31 (.02)\\
GID & .26 (.01) & .43 (.00) & .18 (.01) & .26 (.01) & .44 (.00) & .18 (.00) & \textbf{.34} (.02) & \textbf{.51} (.00) & .24 (.01) & .33 (.01) & \textbf{.51} (.01) & \textbf{.25} (.01)\\
Optimal-31 & .36 (.04) & .64 (.01) & .30 (.02) & .49 (.03) & .74 (.01) & .44 (.03) & .66 (.05) & .92 (.02) & .68 (.05) & \textbf{.72} (.03) & \textbf{.93} (.01) & \textbf{.73} (.03)\\
PatternNet & .90 (.01) & .92 (.00) & .85 (.01) & .94 (.01) & .95 (.00) & .91 (.01) & \textbf{.95} (.01) & \textbf{.96} (.00) & \textbf{.93} (.01) & .94 (.01) & \textbf{.96} (.00) & .92 (.01)\\
RSC11 & .79 (.03) & .82 (.01) & .71 (.02) & .81 (.02) & \textbf{.84} (.01) & \textbf{.74} (.02) & \textbf{.82} (.07) & \textbf{.84} (.02) & .73 (.07) & .79 (.05) & \textbf{.84} (.02) & .73 (.04)\\
RSD46-WHU & .45 (.01) & .58 (.01) & .30 (.01) & .55 (.02) & .67 (.00) & .42 (.02) & .58 (.01) & .67 (.00) & .44 (.01) & \textbf{.63} (.01) & \textbf{.72} (.00) & \textbf{.49} (.01)\\
RSS-CN7 & .46 (.03) & .56 (.01) & .36 (.01) & .53 (.03) & .63 (.01) & .45 (.02) & \textbf{.64} (.06) & .71 (.02) & \textbf{.58} (.06) & .63 (.04) & \textbf{.72} (.02) & .57 (.04)\\
SIRI-WHU & .64 (.02) & .66 (.01) & .48 (.02) & \textbf{.74} (.03) & \textbf{.73} (.01) & \textbf{.60} (.02) & .66 (.03) & .68 (.01) & .56 (.02) & .70 (.01) & .70 (.00) & .58 (.01)\\
WHU-RS19 & .60 (.05) & .82 (.02) & .58 (.04) & .64 (.06) & .84 (.02) & .63 (.05) & \textbf{.76} (.05) & \textbf{.90} (.02) & \textbf{.72} (.05) & .73 (.05) & .89 (.02) & .70 (.06)\\
\midrule
SATIN T2 & .51 & & & .58 & & & .63* & & & \textbf{.64*} &  & \\
\bottomrule
\end{tabular*}
\caption{Average clustering metrics and standard deviations (in parentheses) across 10 replications for the datasets in the SATIN Land Use task (T2). The best results per dataset are highlighted in bold. The SATIN T2 metric denotes the geometric mean of dataset accuracy scores. *We do not include results for the RESISC45 dataset when using a feature extractor finetuned on the same.}
\label{tbl:main}
%\end{table}
\end{sidewaystable}
%\end{landscape}

\subsection{Implementation}
\label{subsec:implement}

All experiments were run on a compute node with 2x Intel Gold 6248 CPUs and a Quadro RTX 8000. For reproducibility, our code is fully containerised using CharlieCloud \citep{priedhorsky_charliecloud_2017} and will be made available. To assess the performance of our proposed method, we use datasets with known ground truth labels to calculate three common clustering metrics: Clustering Accuracy (ACC), Normalised Mutual Information (NMI), and Adjusted Rand Index (ARI). We compute clustering accuracy in the standard way using the Hungarian matching algorithm \citep{kuhn_hungarian_1955}. For all of these metrics, larger numbers indicate better performance. Results presented in tables are rounded to 2 decimal places with dropped leading zeros.

\subsubsection{Finetuning}
To perform finetuning, we first generate a 90/10 training/validation split stratified by label. \citet{naushad_deep_2021} investigated the effect of various data augmentation techniques on the transferability of pretrained CNNs for remote sensing image classification and found that random cropping, horizontal flipping and colour jittering improved the generalisation ability of the models, which we adopt for our own source data preprocessing. In particular, for training images we apply a 10\% jitter to brightness, contrast, and saturation to improve robustness \citep{naseer_intriguing_2021}. Our optimiser was chosen to be similar to the pretraining regime for DINOv2; we use the Transformers library \citep{wolf_transformers_2020} to perform 10 epochs of finetuning with a batch size of 32 using the AdamW optimizer \citep{loshchilov_decoupled_2017} with a weight decay of .01 and accuracy as our metric.

\subsubsection{Manifold Projection}
We use the Uniform Manifold Approximation and Projection (UMAP) algorithm \citep{mcinnes_umap_2018} for all our results. In all cases, we fix the number of neighbours to be 100, the minimum separation distance of embedded points to be 0.5, and the embedding dimension to be $p = 2$. 

\subsubsection{Clustering Algorithm}
We use the Dirichlet Process Gaussian Mixture Model as our clustering algorithm, fit using variational inference (VI-DPGMM) \citep{blei_variational_2006, bishop_pattern_2006_dpgmmvi} as implemented in Scikit-learn \citep{pedregosa_scikit-learn_2011}. We use the same Gaussian-Wishart prior specification and hyperparameter settings for all of our results. Under the stick-breaking representation \citep{sethuraman1994constructive} of the DPGMM, we truncate the number of components to $\overline{K} = 50$ and choose our DP concentration to be $\alpha = 1/50$. We allow each component to have its own general covariance matrix, and set the prior precision $\gamma$ for each component's mean to be .01 to allow for greater variation from the empirical prior mean. For the Wishart precision prior, we choose the least informative degrees of freedom $\nu = p$ and estimate the prior scale matrix $\textbf{D}$ as the empirical precision. We perform up to 200 variational updates for each of five random cluster initialisations using K-means.

%\begin{figure}[tbh]
%  \centering
%   \includegraphics[width=0.95\linewidth]{figures/eurosat_features_combined_vertical.png}
   %\input{figures/pipeline_fig_tikz}
%   \caption{Visualization of the embedded features for the Optimal-31 dataset when using the base DINOv2-L model as a feature extractor (left), and when using the DINOv2-L model finetuned on the RESISC45 dataset (right). The colors and legend denote the ground truth labels. As verified in \cref{tbl:main}, the features from the finetuned model are much easier to cluster.}
%   \label{fig:eurosat-ft}
%\end{figure}

\subsection{SATIN Benchmark: Land Use}

To benchmark the performance of our method in the remote sensing domain, we consider the SATIN metadataset's Land Use task (T2) datasets \citep{roberts_satin_2023}. Clustering these datasets with a single model is a challenging heterogeneous transfer learning task as they vary greatly in their resolution, labels, spatial distribution, image size, and image count. This heterogeneity is summarised in \cref{tbl:satindata} and example images are shown in \cref{fig:satint2}. The overall SATIN metric for the T2 task is the geometric mean of the accuracies across all datasets. We note that while the RESISC45 result is not included in the SATIN metric for feature extractors finetuned on that dataset, interpolating the accuracies achieved by the corresponding size (Base) feature extractors changes the SATIN metric by less than .01 so this difference is negligible. 

\begin{table}[h]
\small
\centering
\setlength\tabcolsep{1pt}
\begin{tabular*}{\linewidth}{@{\extracolsep{\fill}} lllll}
\toprule
Dataset    & $n$    & Resolution (m) & Width (px) & $K$ \\ 
\midrule
RESISC45 \citep{cheng_remote_2017}   & 31500  & 0.2 - 30       & 256              & 45  \\
CLRS \citep{li_clrs_2020}       & 15000  & 0.26 - 8.85    & 256              & 25  \\
EuroSAT \citep{helber_eurosat_2017}    & 27000  & 10             & 64               & 10  \\
GID \citep{GID2020}       & 30000  & 4              & 56 - 224         & 15  \\
Optimal-31 \citep{wang2018scene} & 1860   & 0.2 - 1        & 256              & 31  \\
PatternNet \citep{zhou_patternnet_2018} & 30400  & 0.06 - 4.7     & 256              & 38  \\
RSC11 \citep{zhao2016feature}     & 1232   & 0.2            & 512              & 11  \\
RSD46-WHU \citep{long2017accurate, xiao2017high} & 117000 & 0.5 - 2        & 256              & 46  \\
RSSCN7 \citep{zou_2015_rsscn7}    & 2800   & 1 - 8          & 400              & 7   \\
SIRI-WHU \citep{zhao2015dirichlet, zhao2016fisher, zhu2016bag}  & 2400   & 2              & 200              & 12  \\
WHU-RS19 \citep{xia2009structural, dai2010satellite}  & 1013   & 0.5            & 600              & 19  \\ 
\bottomrule
\end{tabular*}
\caption{Summary of the image count, spatial resolution captured per pixel (ground sampling distance), image size, and number of ground truth labels for datasets in the SATIN Land Use (T2) metadataset. Here, '-' denotes a range of possible values.}
\label{tbl:satindata}
\end{table}

Comparing the results in \cref{tbl:main} across different feature extractors, we make two remarks. First, the larger models generally outperform the base size models. This is an unsurprising result; overwhelmingly it has been found that larger transformer models result in more informative features and better performance in downstream tasks \citep{dosovitskiy_image_2020, oquab_dinov2_2023, adaloglou_exploring_2023, radford_learning_2021, liu_remoteclip_2023, roberts_satin_2023, cong_satmae_2022, li_unsupervised_2022, li_zero-shot_2017}. Second, the models finetuned on RESISC45 generally outperform the Base models. This result is slightly more surprising considering that \citet{oquab_dinov2_2023} found finetuning to be optional when using DINOv2 models. However, it is consistent with recent literature on using finetuned models for heterogeneous transfer learning \citep{dimitrovski_aitlas_2023, dimitrovski_current_2023, naushad_deep_2021, ayush_geography-aware_2020, wang_self-supervised_2022, li_unsupervised_2022}. \cref{fig:optimal31-ft} shows the embedded features for the Optimal-31 dataset with and without finetuning on RESISC45. It demonstrates that the finetuning procedure results in the extracted features that better account for the feature and label shift from the Base model's training images to the target remote sensing images.

\subsection{State-of-the-Art Comparison}

\begin{table}[h]
\centering
%\setlength\tabcolsep{1pt}
%\begin{tabular*}{\linewidth}{@{\extracolsep{\fill}} llll}
\begin{tabular}{llll}
\toprule
Dataset    & Our Method & OpenCLIP & RemoteCLIP \\ 
\midrule
CLRS       & \textbf{.79} & .68          & -            \\
EuroSAT    & .35          & \textbf{.62} & .60          \\
GID        & \textbf{.33} & .31          & -            \\
Optimal-31 & .72          & .84          & \textbf{.90} \\
PatternNet & \textbf{.94} & .80          & .69          \\
RSC11      & \textbf{.79} & .63          & .75          \\
RSD46-WHU  & \textbf{.63} & .44          & -            \\
RSSCN7     & .63          & \textbf{.66} & -            \\
SIRI-WHU   & \textbf{.70} & .56          & -            \\
WHU-RS19   & .73          & .88          & \textbf{.95} \\
\bottomrule
\end{tabular}
\caption{Comparison of the clustering accuracy of our method and the classification accuracy of state-of-the-art zero-shot methods (OpenCLIP, RemoteCLIP) on the SATIN Land Use task (T2). The results shown for our method are when using the DINOv2-L (RESISC45) model as a feature extractor.}
\label{tbl:zeroshot-compare}
\end{table}

Although the SATIN benchmark was originally designed to assess zero-shot classification accuracy using pretrained models, we use it to assess the unsupervised clustering accuracy of our method. In \cref{tbl:zeroshot-compare} we compare the unsupervised accuracy of our method with the classification accuracy achieved by current state-of-the-art zero-shot classification models OpenCLIP \citep{cherti_reproducible_2022, roberts_satin_2023} and RemoteCLIP \citep{liu_remoteclip_2023}. We note that clustering accuracy is a more challenging metric than classification accuracy in our framework due to its sensitivity to the estimated number of clusters, not to mention our method relying only on the vision modality.

\begin{figure*}[tbh]
  \centering
   \includegraphics[width=0.95\linewidth]{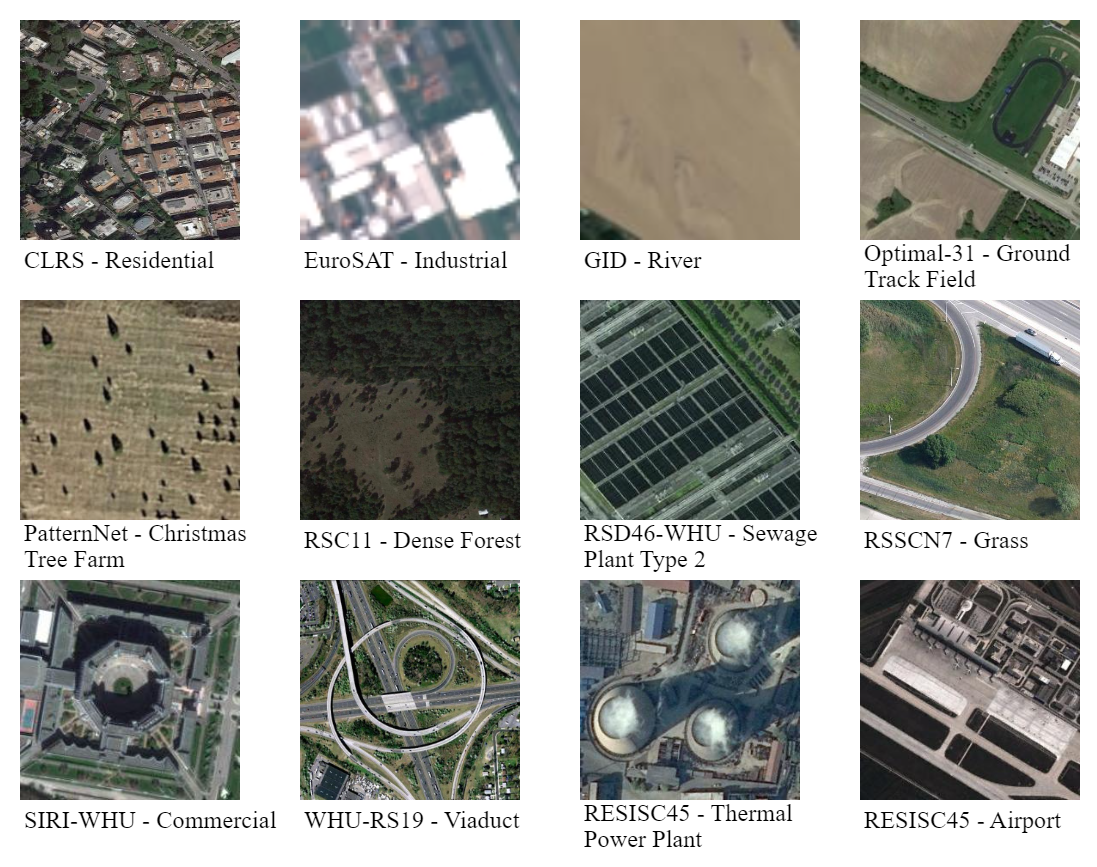}
   \caption{Example images and their ground truth label from each of the SATIN Land Use task (T2) datasets. Aspects such as ground truth labels, image size, and spatial resolution (ground sampling distance) visually appear to be highly heterogeneous across different datasets.}
   \label{fig:satint2}
\end{figure*}

Despite the much larger ViT-G backbone of the OpenCLIP model and the extensive domain-specific training of the RemoteCLIP model, our method gives competitive performance on most datasets in the SATIN Land Use task. Recent literature suggests that this may be due to vision-language models (VLMs) suffering from an imbalance in the information sharing between text and vison modalities; in particular, representations learnt by the text encoder are more influenced by the vision modality than vice versa \citep{frank_vision-and-language_2021, yuksekgonul_when_2023}. However, from both \cref{tbl:main} and \cref{tbl:zeroshot-compare} we note that the performance of our method on the EuroSAT and GID datasets is unimpressive compared to our other results. We believe that this discrepancy is due to the low resolution (ground sampling distance) of these two datasets compared to the other datasets we consider, as described in \cref{tbl:satindata} and shown in \cref{fig:satint2}. Because vision transformer models can achieve excellent classification performance on these datasets \citep{naushad_deep_2021}, we hypothesise that our choice of manifold embedding into only $p=2$ dimensions is not sufficient to capture this extreme scale difference.  We also note that because OpenCLIP and RemoteCLIP are VLMs, their better performance on EuroSAT may be due to the advantage of using both text and imaging modalities compared to our method that uses only imaging \citep{frank_vision-and-language_2021, yuksekgonul_when_2023}. 

\subsection{Alternative Clustering Methods}
\label{subsec:altclust}

Finally, we compare our Bayesian nonparametric clustering approach with two other popular clustering algorithms, HDBSCAN \citep{mcinnes_accelerated_2017} and K-Means \cite{lloyd_least_1982}. Since there are situations in which the true number of cluster labels is known a priori, we examine K-Means in the best case scenario by giving it the true number of clusters. Here, we choose to compare the NMI as it is a better judge of clustering performance across different algorithms than accuracy \citep{rosenberg-hirschberg-2007-v}. For all clustering algorithms, we use the implementation from Scikit-learn \cite{pedregosa_scikit-learn_2011} with default parameters unless otherwise specified. When fitting K-Means, we use the ground truth number of labels and five random initializations. When fitting HDBSCAN, we use a minimum estimated cluster size of 10.

From \cref{tbl:clustering} we find that the clustering performance of the DPGMM is competitive with K-Means despite having to infer the number of clusters from the data using the same prior for every dataset. The poor and highly inconsistent performance of HDBSCAN (as evidenced by the large standard deviations) was surprising to us, but this pattern persisted even after verifying our code and rerunning with various hyperparameter adjustments. In general, its performance was very sensitive to minor changes in the manifold embedding. As such, for practitioners we recommend using the DPGMM in general and K-Means only if the true number of clusters is known a priori.

\begin{table}[h]
\centering
\begin{tabular}{llll}
\toprule
Dataset & DPGMM & HDBSCAN & *K-Means \\
\midrule
CLRS & .78 (.00) & .71 (.01) & \textbf{.79} (.00)\\
EuroSAT & .58 (.01) & .02 (.01) & \textbf{.63} (.01)\\
GID & .51 (.01) & .15 (.18) & \textbf{.52} (.00)\\
Optimal-31 & .93 (.01) & \textbf{.99} (.00) & \textbf{.99} (.00)\\
PatternNet & \textbf{.96} (.00) & .93 (.00) & \textbf{.96} (.00)\\
RSC11 & .84 (.02) & .70 (.09) & \textbf{.87} (.01)\\
RSD46-WHU & \textbf{.72} (.00) & .45 (.25) & .71 (.00)\\
RSSCN7 & \textbf{.72} (.02) & .54 (.10) & .70 (.05)\\
SIRI-WHU & \textbf{.70} (.00) & .38 (.31) & .67 (.01)\\
WHU-RS19 & .89 (.02) & .94 (.01) & \textbf{.96} (.00)\\
\bottomrule
\end{tabular}
\caption{Average NMI and standard deviations (in parentheses) across 10 replications for the datasets in the SATIN Land Use task (T2) when using the DINOv2-L (RESISC45) model as a feature extractor. *Note that K-Means is using the ground truth K.}
\label{tbl:clustering}
\end{table}
\section{Conclusion}
\label{sec:conclusion}

\subsection{Limitations} As demonstrated in \cref{sec:experiment}, the effectiveness of our method is highly dependent on the quality of the extracted features. Although we demonstrated that its performance is relatively robust to keeping hyperparameters fixed across various datasets, it is generally nontrivial to optimise these values in an unsupervised setting. And, as seen in \cref{tbl:clustering}, practitioners who know the number of ground truth labels may benefit from using parametric clustering algorithms such as K-means.

\subsection{Future Extensions} Since neural networks and UMAP are already capable of online inference, an interesting extension to our current work would be to replace the standard VI-DPGMM with an online variant to allow continuous clustering of streamed images \citep{hughes_memoized_2013, bing_diva_2023}. This is particularly relevant for remote sensing applications where hundreds of terabytes of unlabelled images are being produced daily \citep{national_noaa_nodate, crosier_how_2022}. Another interesting direction would be to explore unsupervised finetuning strategies to further relax the mild assumption of a labelled source dataset in the target domain. We would also like to explore manifold embeddings into non-Euclidean spaces, such as the space of hyperspheres \citep{pmlr-v38-straub15, mcinnes_umap_2018}, or directly clustering the estimated weighted graph to avoid having to specify an embedding dimension $p$ \citep{morup_bayesian_2012}. 

\subsection{Summary} We proposed a pipeline for grouping images using neither labels nor a prespecified number of clusters, and implemented it using DINOv2 pretrained visual models, Uniform Manifold Approximation \& Projection, and the Dirichlet Process Gaussian Mixture Model. We applied the method to a variety of different land use classification benchmarks and found that its performance was near or exceeding current state-of-the-art approaches.

\subsubsection{Acknowledgements}
This research used resources provided by the Darwin testbed at Los Alamos National Laboratory (LANL) which is funded by the Computational Systems and Software Environments subprogram of LANL's Advanced Simulation and Computing program (NNSA/DOE). This project was supported by the National Science Foundation's (NSF) Division of Mathematical Sciences.

\subsubsection{Data Availability Statement} Data sharing is not applicable to this article as no new data were created or analysed in this study.

\subsubsection{Disclosure of Interest} The authors report no conflict of interest.

\bibliographystyle{plainnat}
\bibliography{main}

\appendix

\section{Deep Clustering Performance on CIFAR-10}
\subsection{Validation on CIFAR-10}

In this appendix, we present a proof-of-concept to validate the effectiveness of the image clustering pipeline described in the main text. Our goal is to demonstrate that the manifold projection and clustering techniques employed are not only effective for remote sensing, but also appropriate for general clustering tasks in computer vision. To demonstrate that the choice of UMAP and the DPGMM are appropriate, we compare the clustering performance of our pipeline when no finetuning was done with other state-of-the-art image clustering techniques on a general image benchmark.

Because the DINOv2 models included certain benchmark datasets such as STL-10 \citep{coates_analysis_2011} and ImageNet \citep{deng_imagenet_2009} in their training data, we cannot use them for a fair comparison with other clustering methods. Instead, we focus on validating our method on a common clustering benchmark not included in the training data, CIFAR-10 \citep{krizhevsky_learning_2009}. Our method achieves a ACC of .991 when using the DINOv2-L (Base) model as a feature extractor; nearly closing the gap with the state-of-the-art supervised classification accuracy of .995 achieved by \citet{oquab_dinov2_2023}. As shown in \cref{tab:cifar10}, this performance beats current state-of-the-art image clustering methods such as SPICE \citep{niu_spice_2022} and TEMI CLIP ViT-L \citep{adaloglou_exploring_2023} despite not requiring the number of ground truth labels to be specified.

\begin{table}[h]
    \centering
    \begin{tabular}{lccc}
        \toprule
        Method & ACC & NMI & ARI \\
        \midrule
        Our Method & \textbf{.991} & \textbf{.974} & \textbf{.980} \\
        TEMI CLIP ViT-L & .969 & .926 & .932 \\
        SPICE & .918 & .850 & .836 \\
        \bottomrule
    \end{tabular}
    \caption{Comparison of our clustering performance metrics (ACC, NMI, ARI) on the CIFAR-10 dataset against state-of-the-art deep clustering approaches. The results shown for our method are when using the DINOv2-L (Base) model as a feature extractor.}
    \label{tab:cifar10}
\end{table}

This strong performance relative to the classification task with the same feature extractor further reinforces our observation that the most important choice for clustering performance is the feature extractor.

%\processdelayedfloats
%\listoffigures

\end{document}